\documentclass{bmvc2k}

\usepackage{multirow}
\usepackage{amssymb}
\usepackage{footmisc}

\newcommand\blfootnote[1]{%
  \begingroup
  \renewcommand\thefootnote{}\footnote{#1}%
  \addtocounter{footnote}{-1}%
  \endgroup
}

\def\etal{\emph{et al}\bmvaOneDot}

\title{Deep View-Sensitive Pedestrian Attribute Inference in an end-to-end Model}
\addauthor{M. Saquib Sarfraz$^*$}{saquib.sarfraz@kit.edu}{1}
\addauthor{Arne Schumann$^*$}{arne.schumann@iosb.fraunhofer.de}{2}
\addauthor{Yan Wang}{uudqa@student.kit.edu}{1}
\addauthor{Rainer Stiefelhagen}{rainer.stiefelhagen@kit.edu}{1}
\addinstitution{
Institute of Anthropomatics \& Robotics\\
Karlsruhe Institute of Technology\\
Karlsruhe, Germany
}
\addinstitution{
Fraunhofer IOSB\\
Fraunhoferstr. 1,\\
Karlsruhe, Germany
}
\runninghead{Sarfraz \MakeLowercase{\etal}}{View-Sensitive Pedestrian Attribute Inference}

\begin{document}
\maketitle
\blfootnote{$^*$Arne Schumann and M. Saquib Sarfraz contributed equally to this work.}

\begin{abstract}
Pedestrian attribute inference is a demanding problem in visual surveillance that can facilitate
person retrieval, search and indexing. To exploit semantic relations between attributes,
recent research treats it as a multi-label image classification task. 
The visual cues hinting at attributes can be strongly localized and inference of person attributes
such as hair, backpack, shorts, etc., are highly dependent on the acquired view of the pedestrian. In
this paper we assert this dependence in an end-to-end learning framework and show that a
view-sensitive attribute inference is able to learn better attribute predictions. Our proposed model
jointly predicts the coarse pose (view) of the pedestrian and learns specialized view-specific
multi-label attribute predictions. We show in an extensive evaluation on three challenging
datasets (PETA, RAP and WIDER) that our proposed end-to-end view-aware attribute
prediction model provides competitive performance and improves on the published state-of-the-art on
these datasets.
\end{abstract}

\section{Introduction}
\label{sec:intro}

Person attribute recognition in surveillance footage is a highly demanding problem as it benefits
several related applications such as image indexing, person retrieval
\cite{siddiquie2011image,feris2014attribute} and person re-identification \cite{layne2012person}.
Methods addressing pedestrian attribute recognition in such applications have to deal with
challenges due to low resolution, detected pedestrians in far-range surveillance scenes, pose
variations, and occlusions. The task is to make predictions for a set of attributes
given an image of a person as input. In contrast to the general image recognition problem where each
image has one label of a certain class, the pedestrian attribute inference is a multi-label
recognition problem where each of the pedestrian images is assigned a multitude of semantic attribute
labels with binary outcome, e.g., wearing short skirt, male, running, carrying backpack, etc.

\begin{figure*}[th!]
    \begin{center}
        \includegraphics[width=12cm,height=4cm]{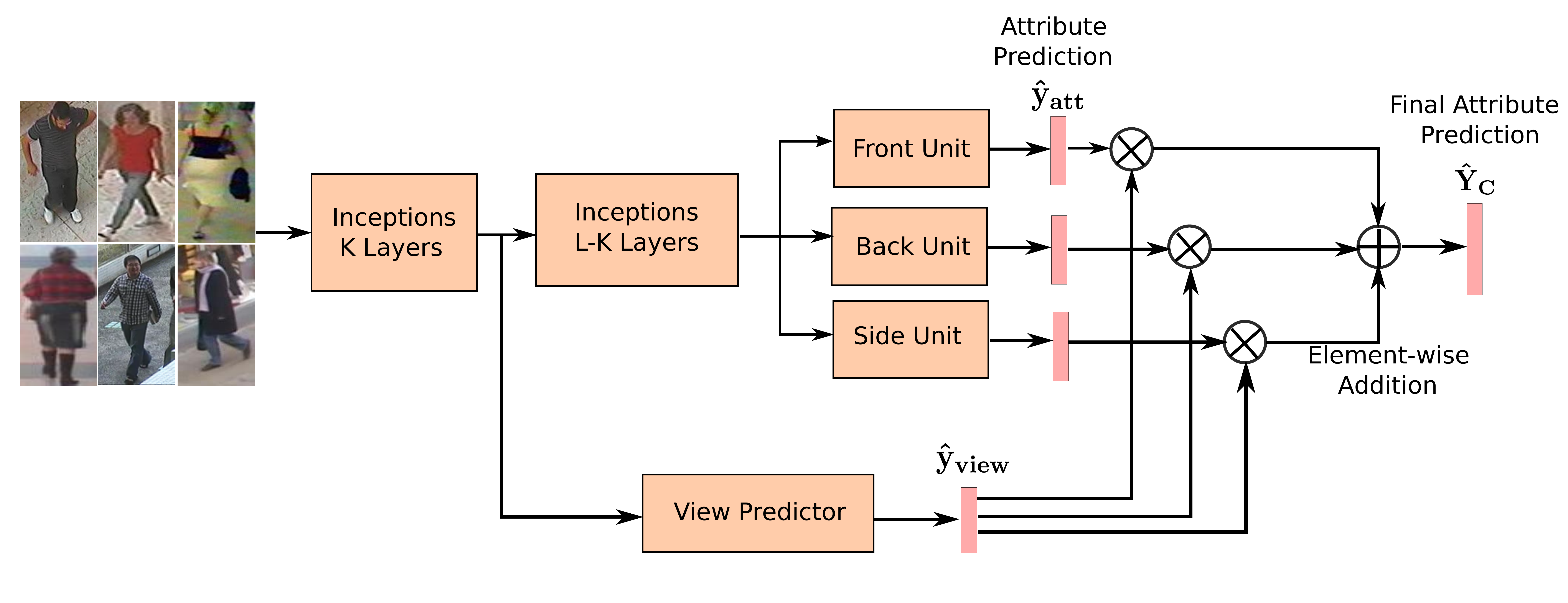}
    \end{center}
    \caption{Overview of our View-Sensitive Pedestrian Attribute (VeSPA) model.}
    \label{fig:overview}
\end{figure*}
% * <saquibsarfraz@gmail.com> 2017-05-02T09:57:04.572Z:
%
% > \begin{figure*}[th!]
% >     \begin{center}
% >         \includegraphics[width=12cm,height=4cm]{images/intro}
% >     \end{center}
% >     \caption{Challenging attribute recognition examples. Some attributes are hard to recognize in 
% >     certain views. Our view specialized model successfully recognizes these attributes.}
% >     \label{fig:intro}
% > \end{figure*}
%
% ^.

Most previous methods solve the multi-label person attribute recognition by optimizing a separate
binary classification model for each of the attributes
\cite{panda2014,li2015multi,sudowe2015}. Such a direction, however, ignores the semantic
relationships between attributes. To better exploit these semantic relationships more recent
proposals have shown improved performance by solving the task as a direct multi-label classification
problem \cite{li2016richly,zhu2017,li2016human}. 
While multi-label classification generally achieves better accuracy than binary classifiaction 
approaches, few approaches explicitly capture the spatial
relationships of attributes (i.e. the location of attributes in the image). This can have several
associated problems as each labeled attribute may intrinsically be tied to very different image
regions based on the acquired view of the pedestrian. As an example, carrying backpack
in a back-view image has a very different and much larger spatial context to learn from in 
comparison with a side- or front-view. If this spatial context is known, it may better
guide the training process to focus on respective image regions for each attribute.
In a very recent proposal Zhu et al. \cite{zhu2017} try to infer this localization explicitly in the
model and have shown improved performance.
While spatial context is relevant for some of the attributes it is not pertinent for other, more 
global attributes such as gender, action, age, etc.

Another solution is to guide the learning process by using additional part or pose
information. As in the example above, the spatial context is also tied to the 
pose of the person. The pose information may also provide better context in
explaining some of the global level attributes like gender, age, or action.  Some recent methods
\cite{panda2014} \cite{li2016human} proposed to learn separate models depending upon \textit{a
priori} selected parts, pose and context. In this paper we propose to learn an end-to-end unified
model to jointly learn the coarse pose prediction (front-, back-, or side-view) and specialized,
view-dependent attribute inference in a multi-label classification setting. We use a deep
Convolutional Neural Network (CNN) to train a view predictor in the early layers and
separate view-specific attribute prediction units in the later layers of the same model.
Figure \ref{fig:overview} depicts an overview of our View-Sensitive Pedestrian Attribute approach
(VeSPA).

Our main
contribution is to show that in addition to the popular view of relying on either body parts, attribute
spatial context in the image, or general scene context, the coarse body pose (view) information can be
another simple yet highly relevant clue for reliable attributes inference. Our results
show that different attributes relate differently to the acquired view of the person and learning
the views helps the overall attribute prediction more. Our evaluation on three of the largest
available datasets, the PETA and RAP surveillance datasets, and the WIDER attribute dataset which
features challenging person images in photos, shows convincing results.

In the remainder of this work we first discuss related approaches in Section \ref{sec:relwork}.
Section \ref{sec:mth1} provides details of the proposed approach. We conclude in Section
\ref{sec:conclusion} after discussing the results of our comprehensive evaluations in Section
\ref{sec:eval}.

\section{Related Work}
\label{sec:relwork}
We limit our discussion of related work to attribute recognition approaches which pertain to person
or pedestrian images. 

% general approaches -- single label
%\noindent\textbf{General Attribute Recognition}
Attribute classification is a multi-label classification task. A straightforward way to address this
is by relying on the extensive single-label classification literature and training a separate
classifier for each attribute.
Sharma et al. \cite{sharma2011learning} apply this approach by using spatial histogram features in 
conjunction with a maximum margin optimization to learn each of the attribute classes. 
Similarly, Layne et al. \cite{layne2012person} and Deng et al. \cite{deng2015learning} use Support 
Vector Machines (SVMs) and a set of color and texture features to classify each attribute.
Such approaches cannot directly leverage semantic relations between
attributes.
%
%
%
% general approaches - multi-label
%
More recent approaches often rely on a single model to recognize all
attributes (multi-label classification).
Sudowe et al. \cite{sudowe2015} describe a person attribute CNN which is trained with one loss
for each attribute. The main part of the net is shared for all attributes and allows the approach to
implicitly leverage attribute relationship information.
A similar approach with individual attribute losses which are manually restricted to relevant body
parts is described by Zhu et al. in \cite{zhu2017multi}.
In \cite{li2015multi} Li et al. show that it can be beneficial to train an attribute CNN with a 
single, weighted loss which includes all attributes and applies weights based on each attribute's 
label imbalance.
This approach is extended by Li et al. in \cite{li2016richly} into a part-based model.
In \cite{yu2016weakly} Yu et al. propose a CNN based approach which relies on multi-level 
deep features and is able to recognize as well as localize pedestrian attributes. This approach also 
relies on a single, weighted loss.
Joint recognition and localization of general image attributes is also performed in \cite{zhu2017}
and applied to person attribute recognition.

%\noindent\textbf{Pose-aware Attribute Recognition}
The importance of person pose information for the attribute recognition task has been studied in 
several works.
In \cite{bourdev2011describing} Bourdev et al. use pose-sensitive body part detectors and apply
attribute classifiers for each part detection. Each attribute classifier is thus specific to a
certain body part and pose. A main drawback is the large number of resulting attribute classifiers.
In a later work \cite{panda2014} the same part detectors are used to generate a pose-normalized 
person representation based on deep features which is used for attribute recognition with linear 
SVMs.
Park et al. \cite{park2016attribute} describe a deep model which jointly learns to detect person
keypoints, body parts and attributes. Pose information is implicitly contained in normalized body
part representation and attributes are manually assigned to the relevant body parts.
In \cite{yang2016attribute} Yang et al. learn a joint model for body part localization and attribute
recognition which detects keypoints and generates a warping matrix for pose normalization. 
Another recent approach by Li et al. \cite{li2016human} relies on full image and leverages body 
parts and scene context information to more accurately determine person attributes in a combined 
deep model.
Most of these approaches aim to include additional information, localization (explicitly or by a 
part-based approach) or context knowledge, to aid in the attribute recognition task. However, they 
do not explicitly rely on a person's acquired view which our experiments show 
is a crucial clue for robust attribute recognition.
%
% mention that coarser is more robustly detected/classified? -> have no results that show this

\section{View-Sensitive Pedestrian Attribute Inference}
\label{sec:mth1}
We adapt a deep neural network for joint pose and multi-label attribute classification. The overall
design of our approach is shown in Figure \ref{fig:overview}. The main network is based on the
GoogleNet inception architecture \cite{szegedy2015}. As shown, the network contains a
view classification branch and three view-specific attribute predictor units. The view classifier
and attribute predictors are both trained with separate loss functions. Prediction scores from
weighted view-specific predictors are aggregated to generate the final multi-class attribute
predictions. The whole network is a unified framework and is trained in an end-to-end manner.

\subsection{End-to-end View \& View-aware Attribute Classification}
\label{subsec:mth1}

Let \textbf{I} denote an input person image with ground-truth labels $Y=\{Y_{C};Y_{V}\}$.
$ Y_{C} = [y^{1}, y^{2},..., y^{C}]^{T} $ denotes the attribute labels, where $y^{i}$ is a binary
indicator: $y^{i} = 1$, if image \textbf{I} is tagged with attribute $i$ and $y^{i}=0$ otherwise. 
$C$ is the number of attributes in the dataset.
The view label is denoted by $Y_{V}\in\{front, back, side\} $. The overall network conducts
view prediction and multi-label attribute inference using two different corresponding losses.

The main net is based on the GoogleNet inception architecture. It has repetitive inception blocks
where each inception module ranges from 256 filters in the early modules to 1024 in top inception
modules. Our design shares the same network for both tasks. The coarse pose or view prediction is
based on the output of early layers. The view predictor is a branch out after the $K$-th layer of
the model and conducts classification for each of the three views $Y_{V}$ of the image \textbf{I}:
\begin{align}
    \mathbf{X}_{K}&=f_{K}(\mathbf{I};\theta_{K}),\;\; \mathbf{X}_{K} \in\mathbb{R}^{n \times n \times
    k}\nonumber
    \\
    \mathbf{\hat{y}}_{view}&=f_{view}(\mathbf{X}_{K};\theta_{view}),\;\;%\\
    \mathbf{\hat{y}}_{view} \in \mathbb{R}^{3}
    %, 0 \leq \mathbf{\hat{y}}_{view}(i) \leq 1.\nonumber
\label{eq1}
\end{align}
%\note{$\mathbb{R}^{V}$ is too general, it's a probability vector...}
%\note{Also $I$ here fat, in text not...}
\\ Here, $\mathbf{X}_{K}$ are the $k$ feature maps of size $n \times n$ from layer $K$ and
$\mathbf{\hat{y}}_{view}=[\hat{y}_{view}^{1},\cdots ,\hat{y}_{view}^{V}]^{T}$ are the 
confidences of the view-predictor. These confidences act as weights for the output of the 
corresponding view-specific attribute inference units.

The attribute inference is carried out by branching out separate CNN units, one for each view. The
input to each view-specific unit is the output feature map $\mathbf{X}_{L-1}$ of top level layer $L-1$. The output of
each view-specific unit is an attribute prediction 
$\mathbf{ \hat{y}}_{att} = [y^{1}, y^{2},..., y^{c}]^{T} $ for all the $C$ attributes classes:
\begin{align}
	\mathbf{X}_{L-1}&=f_{L-1}(\mathbf{I};\theta_{L-1}),\;\; \mathbf{X}_{L-1} \in\mathbb{R}^{n \times n \times k}\nonumber
    \\
	\mathbf{\hat{y}}_{att}&=f_{att}(\mathbf{X}_{L-1};\Theta),\;\;
    \mathbf{\hat{y}}_{att} \in \mathbb{R}^{C}
\label{eq2}
\end{align}
\\ Here, $\Theta=\{\theta_{L-1};\theta_{view}\}$ are the model parameters of the whole net. The
view-specific attribute predictions are weighted by the corresponding predicted view-confidence
$\mathbf{\hat{y}}_{view}$.
%The weighing is an element-wise multiplication of the view-confidence with the attribute predictions.
The final multi-class attribute prediction $\mathbf{\hat{Y}_{C}}=
[y^{1}, y^{2},..., y^{C}]^{T}$ are the aggregated predictions of these weighted view-specific
predictions:

\begin{align}
    \mathbf{\hat{Y}_{C}}= \sum_{V}\mathbf{\hat{y}}_{att}^{V} \circ \mathbf{\hat{y}}_{view}^{V}
\label{eq3}
\end{align}
Note that the weighting by view predictor confidence not only weights the output of the attribute
unit but importantly also weights its gradient equivalently. Thus, a unit whose corresponding view prediction 
weight is low will only receive very small parameter updates for the current training sample and 
thus focus its learning mainly on those samples that receive a correspondingly high confidence for the
given view.
With this design each of the three view-specific units is specialized in inferring attributes from
the respective view and the final aggregation helps share the strengths of each of the units to
improve the final prediction on all images. The whole network is trained with two losses. We use a
3-way softmax for the view-predictor branch. To cast the problem as a multi-label
classification, following \cite{li2015multi}, we use a modified weighted cross-entropy loss at the
final attribute inference layer:
\begin{align}
    L_{attr}= -\dfrac{1}{N}\sum_{i=1}^{N}\sum_{c=1}^{C}w_{c}(y_{ic})log(\hat{y}_{ic}) + (1-y_{ic})log(1-\hat{y}_{ic})
\end{align}
where $w_{c}=exp(-a_{c})$ is the weight for $c$-th attribute. $a_{c}$ is the prior distribution of
the $c$-th attribute in the training set. This is important, because of the large imbalance of
attribute label values in the dataset. $\hat{y}_{ic}$ is the estimated probability for the $c$-th
attribute of the image \textbf{I}.
During training we prevent the gradient of the multi-label attribute from flowing back through the
pose predictor branch. Both, the view predictor gradient and
attribute gradient flow back through the earlier $K$-layers which are updated by the sum of
these two gradients.
\\
\\
\textbf{Implementation Details:} 
The standard GoogleNet inception architecture has two auxiliary loss layers that serve to strengthen
the gradient and encourage discrimination in the lower stages of the network. In our design, we adapt one of
these auxiliary loss layers as our view predictor and remove the other. The view predictor takes as input the
$576$ feature maps from the \emph{inception 3c} layer of the net and comprises of a 5x5 max pooling,
1x1 conv block and two fully connected \textit{fc} layers, where the last \textit{fc} is the
3-dimensional input to the softmax to 
predict the view of the input pedestrian image. Each of the view-specific attribute
units is a separate inception block which takes as input the $1024$ feature maps of the model's
\emph{inception 5a}
layer. Each unit contains 4 parallel strands which use chained convolutional blocks or pooling to
achieve views of varying receptive fields on the input data. The output of each unit is
concatenated, pooled and fed into a \textit{fc} layer of $C$ dimensions as the final output of the unit.
%1, 2, 3 convolutional blocks for the first 3 rows in sequence, and
%a 5x5 max pooling block followed by a 1x1 conv block for the last row. The first conv blocks in the
%first 3 rows are 1x1 conv and the rest conv blocks are 3x3. Finally, the outputs of all 4 rows (352,
%320, 224, 128 feature maps in sequence of rows) are concatenated together as the output of each
%unit. 
The output of all three units are weighted by the view-predictor by 
multiplication and aggregated together by element-wise addition before being finally fed to the
attribute loss.

To avoid overfitting and achieve a more robust attribute recognition, we apply image augmentation
during training. Images are first normalized to zero-mean and resized to $256 \times 256$.
For batch creation we then randomly crop the images to the GoogleNet input size of $227 \times 227$
and apply random horizontal flipping.
The network's weights are initialized from a pre-trained ImageNet model and fine-tuned with an
initial learning rate of 0.0002 using the Adam solver with a batch size of 32.
%For testing, we resize all images to $256 \times 256$ and conduct a single center crop evaluation.

\newcommand{\f}[1]{\textbf{#1}} % for top result
\newcommand{\np}[0]{$^\dag$}

\section{Evaluation}
\label{sec:eval}

We evaluate our VeSPA model on three public datasets.
The \textbf{PETA dataset} \cite{deng2014pedestrian} is a collection of several person surveillance 
datasets and consists of 19,000 cropped images. Each image is annotated with 61 binary and 5 
multi-value attributes. Following the established protocol, we limit our experiments to those 35
attributes for which the ratio of positive labels is higher than 5\%. The dataset is 
sampled into 9,500 training images, 1,900 validation images and 7,600 test images.
The \textbf{RAP dataset} \cite{li2016richly} consists of 41,585 person images recorded by
surveillance cameras. Each image is annotated with 72 attributes, viewpoints, occlusions and body
parts. According to the official protocol only those 51 attributes with a positive label ratio above
1\% are used. For our evaluations we split the dataset randomly into 33,268 training images and
8,317 test images.
In order to get a better impression of the performance of our model on data with more complex pose
variation, we also evaluate on the \textbf{WIDER dataset} \cite{li2016human}. The dataset contains
13,789 full images with 57,524 person bounding boxes. 14 attributes are annotated for each person.
We follow the evaluation protocol proposed in \cite{zhu2017} and crop out all bounding boxes. This
results in 28,340 person images for training and validation and 29,177 images for testing. We use all 14
attributes for our experiments on WIDER. The \emph{unspecified} labels of the WIDER dataset are treated as
negative during training and are excluded from evaluation in testing following the settings in \cite{li2016human} \cite{zhu2017}.

Of these three datasets only RAP contains view annotations and allows us to train the view predictor
part of VeSPA. For training on WIDER and PETA we transfer the model learned on the RAP training data
and fix the learning rate of the view predictor at $0$ while training the remainder of the network
as usual.

For our evaluations on PETA and RAP we rely on two types of metrics.
For a \emph{label-based} evaluation we compute the mean accuracy (mA) as the mean of the accuracy
among positive examples and the accuracy among negative examples of an attribute. This metric is not
affected by class imbalances and thus penalizes errors made for the less and more frequent label
value equally strongly. However, this metric does not account for attribute relationships (i.e.
consistency among attributes for a given person example).
In order to account for attribute predictions which are consistent within each person image, we
further use \emph{example-based} metrics. For this we apply the well known metrics accuracy,
precision, recall and F1 score averaged across all examples in the test data.
A more detailed description of the metrics can be found in \cite{li2016richly}.

\subsection{Comparison with State-of-the-art}

\begin{table}[t]
    \begin{center}       
    \resizebox{\columnwidth}{!}{
    \begin{tabular}{|l|c|c|c|c|c||c|c|c|c|c|}
    \hline
    \multirow{2}{*}{Method} & \multicolumn{5}{c||}{RAP}  & \multicolumn{5}{c|}{PETA}   \\
    \cline{2-11}
                            & mA & Acc & Prec & Rec & F1 & mA & Acc & Prec & Rec & F1  \\
    \hline
    ACN \cite{sudowe2015}        & 69.66   & 62.61   &\f{80.12}& 72.26   & 75.98   & 81.15   & 73.66   & 84.06   & 81.26   & 82.64   \\
    DeepMAR \cite{li2015multi}         & 73.79   & 62.02   & 74.92   & 76.21   & 75.56   & 82.89   & 75.07   & 83.68   & 83.14   & 83.41   \\
    DeepMAR$^*$ \cite{li2016richly}\np & 74.44   & 63.67   & 76.53   & 77.47   & 77.00 &-&-&-&-&-\\
    WPAL-GMP \cite{yu2016weakly}\np    &\f{81.25}& 50.30   & 57.17   & 78.39   & 66.12   &\f{85.50}& 76.98&84.07&\f{85.78}&84.90\\
    WPAL-FSPP \cite{yu2016weakly}\np   & 79.48   & 53.30   & 60.82   & 78.80   & 68.65   & 84.16   & 74.62   & 82.66   & 85.16   & 83.40   \\
    \hline
    \hline
    GoogleNet Baseline & 70.11  & 60.88  & 76.62  & 72.65  & 74.58  & 81.98  &76.06  & 84.78  &83.97  & 84.37 \\
       % Ours VeSPA (0.4)          & 79.11   &\f{66.97} & 75.50 & \f{84.05}  &\f{79.54}  & 82.69  &\f{77.47} &\f{85.75}  & 85.22  &\f{85.48}  \\
       % VeSPA (no WPAL) &\f{79.11}&\f{66.97} & 75.50 & \f{84.05}  &\f{79.54}  & 82.69 &\f{77.47}&\f{85.75}  &\f{85.22} &\f{85.48}  \\
       %Ours VeSPA(0.5)           & 77.70   &\f{67.35} & 79.51 & \f{79.67}  &\f{79.59}  & 82.15  &\f{77.21} &\f{86.82}  & 83.82  &\f{85.29}  \\
       %Ours VeSPA (tune from googlenet)           & 77.60   &\f{66.73} & 79.01 & \f{79.08}  &\f{79.04}  & 83.45  &\f{77.73} &\f{86.18}  & 84.81  &\f{85.49}  \\
       Ours VeSPA           & 77.70   &\f{67.35} & 79.51 & \f{79.67}  &\f{79.59}  & 83.45  &\f{77.73} &\f{86.18}  & 84.81  &\f{85.49}  \\
    \hline
    \end{tabular}
    }
    \end{center}
    \caption{Results of our approach on the RAP and PETA datasets. We outperform the
    state-of-the-art on most of the example based metrics. Our strong F1 score indicates a better
    tradeoff between precision and recall than other works (Unpublished works are marked with \np).
      }
    \label{tab:sota-peta-rap}
\end{table}
\begin{table}[t]
    \begin{center}       
    
    \begin{tabular}{|l|c|c|c|c|c|}
    \hline
    Method & RCNN \cite{girshick2015fast}  & R*CNN \cite{gkioxari2015contextual}  & DHC \cite{li2016human}  & ResNet-SRN \cite{zhu2017}  & ours VeSPA   \\
    \hline
    mAP & 80.0  & 80.5 & 81.3 & 86.2 & 82.4 \\
    
    \hline
    \end{tabular}

    \end{center}
    \caption{Results of our approach on the WIDER dataset. VeSPA achieves competitive performance in spite of the much stronger pose
    variation on this dataset.}
    \label{tab:sota-wider}
\end{table}
We compare the performance of our VeSPA model with a number of recent state-of-the-art pedestrian 
attribute recognition works, including ACN \cite{sudowe2015}, DeepMAR \cite{li2015multi}, 
DeepMAR$^*$ \cite{li2016richly} and WPAL \cite{yu2016weakly}. Results of our
approach in context of these works are given in Table \ref{tab:sota-peta-rap}. We also include results of our GoogleNet baseline (without view-units) to demonstrate the gain of the proposed architecture. As seen, the model with view-units performs better. This benefit is more clearly demonstrated on the larger (with more attributes) RAP dataset, where a gain of 4-7\% is achieved across all metrics, with using the specialized view-units. The gain is less pronounced on the PETA dataset (on average 1-2\%).

Our approach achieves competitive performance across all metrics and state-of-the-art results on
some of them. We strongly outperform most other approaches on the example based metrics. Particularly
on accuracy and F1 our approach yields notable improvements over the previous state-of-the-art. The
strong F1 values indicate a good precision-recall tradeoff of our approach.
On both datasets the label based mean accuracy (mA) is lower than that of the unpublished WPAL
approach. However, the example based results of WPAL are much lower in comparison. We have observed
a similar trend during training of our approach. Prolonged training with possible overfitting will
increase the mA metric further at the cost of all example based metrics. We
argue that the example based metrics are more relevant to real world applications as they measure 
the consistency of an attribute-based description of a person which is of greater importance for
communicating such descriptions to security personnel. Furthermore, description consistency is also
important for subsequent tasks, such as person re-identification.

Both, PETA and RAP are typical surveillance datasets (i.e. pedestrian attribute
recognition). While they offer great variety in view angle,
they do not contain a very high degree of person body pose variation.
In order to judge the accuracy of VeSPA under stronger/unknown pose variations, we further compare our
performance to the state-of-the-art on the WIDER dataset (i.e. person attribute recognition).
We compare our performance to R-CNN \cite{girshick2015fast}, R*CNN \cite{gkioxari2015contextual},
DHC \cite{li2016human} and the recent ResNet-SRN \cite{zhu2017}. Results are shown in Table \ref{tab:sota-wider}.
Our approach outperforms the published state-of-the-art, i.e. R-CNN, R*CNN and DHC by at least 1.1\% in mean average precision (mAP).
Interestingly, two of these methods (R*CNN and DHC) make explicit use of scene context and image parts to increase
person attribute precision. Our results show that view information might be a more relevant clue for
attribute recognition than context. The most recent approach, ResNet-SRN \cite{zhu2017} which
simultaneously recognizes and localizes attributes and uses the image-level context of each
attribute outperforms VeSPA in mAP. However, the ResNet-SRN model is highly optimized to learn
the image-level context cues from the WIDER training set whereas our view prediction is merely
transfered from the RAP dataset and no adaptation to views or poses present in WIDER is learned.

Our approach shows competitive or state-of-the-art performance across the three
datasets. Compared to the previous published state-of-the-art, VeSPA has a superior precision-recall trade-off
which makes it particularly suitable for applications that rely on an accurate and consistent
attribute-based description of persons.
%\footnote{We provide more detailed results of VeSPAs performance for individual attributes in the supplementary material.}.

\subsection{View Prediction Analysis}

We investigate the effect of view prediction in aiding the overall attribute inference.  We first provide some quantitative and qualitative performance of the view-classification on the three datasets. RAP is the only dataset which contains view annotations and thus allows for a quantitative evaluation.
%The results of our view predictor subnet on the RAP testset are shown in Table
%\ref{tab:angle-quantitative}. 
Our model achieves a very reliable view classification accuracy of
91.7\%, 91.0\% and 81.3\% for front-view, back-view and side-view, respectively on RAP test set .
This high accuracy is crucial to our approach as it allows the specialized attribute
units of our model to reliably learn view specific information.

%\begin{table}[ht]
%    \begin{center}           
%    \begin{tabular}{|l|c|c|c|}
%    \hline
%    Angle  & Front  & Back  & Side  \\
%    \hline
%    Accuracy  & 91.65  & 91.00  & 81.27  \\
%    \hline
%    \end{tabular}
%    \end{center}
%    \caption{The VeSPA view subnet achieves very reliable predictions of the view angle on the RAP
%    testset. \todo{maybe remove this table, takes up space}}
%    \label{tab:angle-quantitative}
%\end{table}

\begin{figure}[t]
\begin{center}
        \includegraphics[width=\textwidth]{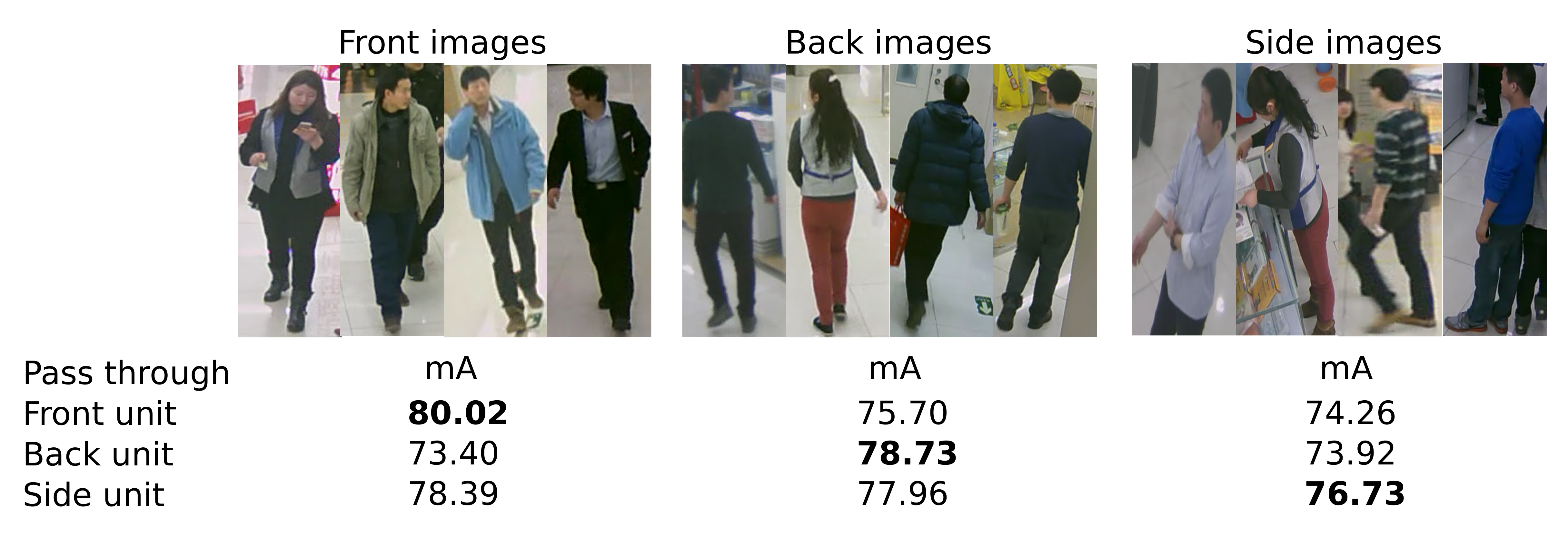}
    \end{center}
    \caption{Specialization of each view-unit:  Attributes mA is calculated by passing the subset of test images belonging to one of the three views through each of the three individual view units, respectively. The bold numbers denote the best performances that are obtained when the images of a specific view are processed by the matching view unit.}
    \label{fig:view-unit}
\end{figure}

In order to quantitatively analyze if the trained view units are indeed specialized to the respective view, we divide the RAP test set according to the three view annotations and tested images of each view separately with all three view-units. For each test, the angle predictor and the other two view-units in the VeSPA model are deactivated. The results are displayed in Figure \ref{fig:view-unit}. The resulting attribute classification accuracy is always the highest for the respective matching view-unit. This shows that our architecture leads to a successful specialization of the view-units.
%The results show at least a 3\% gap between the matching unit and the least accurate unit, e.g. the attribute mA of front images processed by the matching front unit is 6.62\% better than that of the back unit.

\begin{figure}[t]
    \begin{center}
        \includegraphics[width=\textwidth]{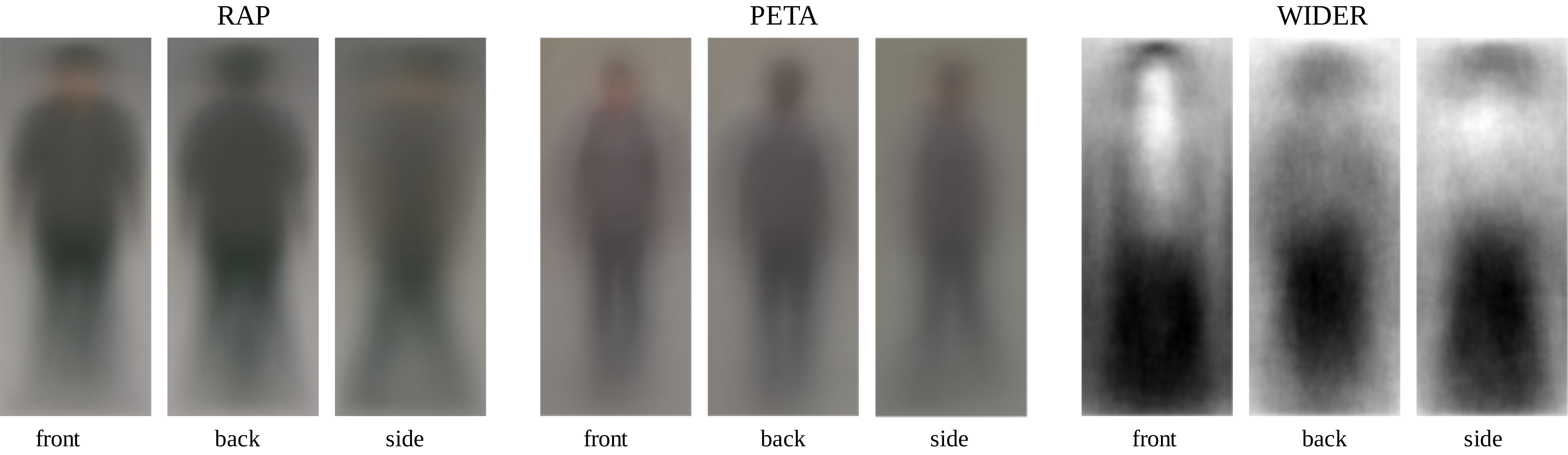}
    \end{center}
    \caption{Mean images of VeSPAs view classification on the test sets of RAP, PETA and WIDER. \textbf{PETA:} Total test images=7600; predicted front (2414), back (2481), side (2705). \textbf{WIDER:} Total test images=29177; predicted front (7216), back (10774), side (11187). Best viewed in color and on screen }
    \label{fig:angle-qualitative}
\end{figure}

\begin{figure*}[t]
    \begin{center}
        \includegraphics[width=\textwidth]{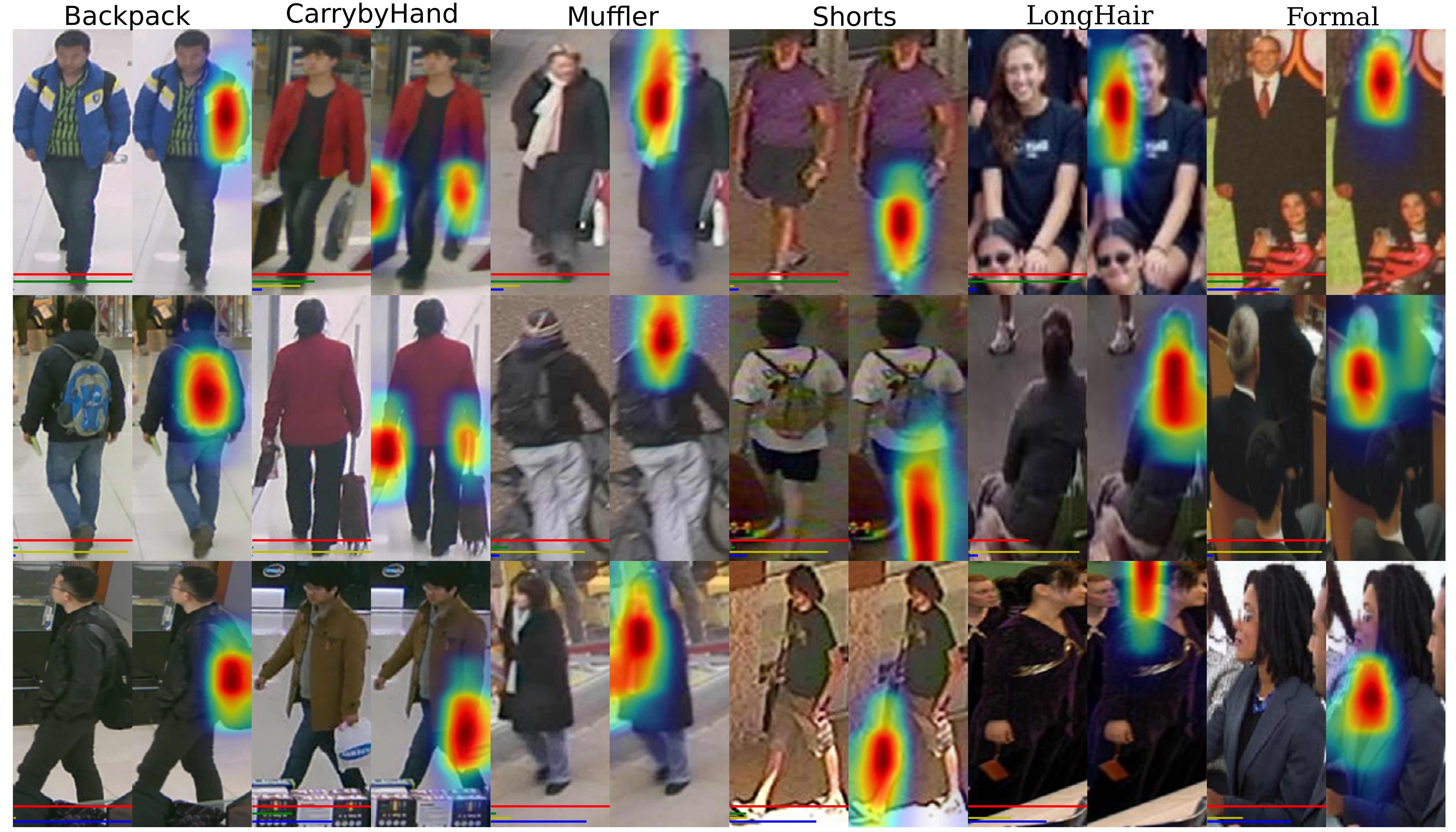}
    \end{center}
    %\caption{Regions most relevant to VeSPAs attribute predictions. The confidence for a positive attribute value is plotted in red. A darker tone of red indicates that blocking the corresponding image region results in a drop of confidence for the attribute prediction (i.e. indicates a region which has high relevance towards the prediction result). The original confidence of VeSPA for the given attribute is plotted as a red bar and the view confidences as green, yellow, and blue bars for front, back, and side respectively on the pedestrian image. Bottom right image for each attribute shows an example where that attribute is not present. (Best viewed in color)}
    \caption{Regions most relevant to VeSPAs attribute predictions: Excitation backprop \cite{zhang2016top} is used to obtain the region localizations on which our model bases its respective attribute prediction. The original confidence of VeSPA for the given attribute is also plotted on the bottom of the images as a red bar and the view confidences as green, yellow, and blue bars for front, back, and side, respectively. Each row of images corresponds to a specific view. Each column shows some representative attribute images from one of the three datasets (RAP, PETA, WIDER). This figure is best viewed in color.}
    \label{fig:relevant-regions}
\end{figure*}

To gain insight into the effects of transferring the view predictor
when training VeSPA on datasets which do not contain view annotations, we provide a qualitative impression in Figure \ref{fig:angle-qualitative} by
computing the mean image for each predicted view across all images on the PETA and WIDER test sets. We compare these mean images to the RAP mean images which we know from our
quantitative analysis to represent very accurate view predictions. The figure shows a very high
resemblance between the PETA and RAP mean images. This indicates a similarly high view
classification accuracy on PETA as on RAP. The three views can be clearly identified by looking at
the images. The side view images are slightly more ambiguous, because left-side and right-side are
not differentiated by our model. WIDER dataset consists of
photos with a much higher degree of background variability, clutter and person pose. This, and a large number of test images (29177) than those of PETA and RAP, leads to an increased blur of the mean image. To make the WIDER image viewable we only show the luminance channel of the color mean image. As seen, in comparison between back-view and front-view a lighter
facial region is still clearly discernible on WIDER.  

We also analyze which image regions are the most relevant to VeSPA's
prediction of a given attribute. To that end, the excitation back-propagation method proposed by \cite{zhang2016top} is applied to our model to generate the attention maps for different attributes. Some of the representative results are shown in Figure \ref{fig:relevant-regions} 
\footnote{See more analysis of such relevant image regions for additional attributes in the supplementary material.}.
%To that end we applied a blocking patch of all zero values at regularly spaced locations in the image and recorded VeSPAs prediction confidence for the given attribute, see figure \ref{fig:relevant-regions}. Strong changes in prediction confidence for a given location of the zero-path indicate an image region that has strong influence on the prediction. We plot the confidences for each patch location with bright red indicating high confidence and darker red indicating a lower confidence for the presence of the attribute. Thus, a mostly red confidence map indicates a prediction for the attribute to be present. Dark areas in this map  indicate locations that contributed most towards this decision. Bottom right image for each attribute in figure \ref{fig:relevant-regions} shows the example where the respective attribute is not present. Conversely the confidence map only gets high when a relevant region is blocked leading the model to predict otherwise.
The images show that, importantly, VeSPA has been able to successfully identify different relevant image regions for the same attribute, under varying views, even though no localization is explicitly included in the training process. For example, the relevant region for the presence of the attribute muffler is the neck area of a person in back-view images but the torso region in side-view images. Some interesting insights are on the clues that are considered most relevant for an attribute which has no well defined localized appearance in the image, e.g. the most relevant clue for the attribute \textit{formal} appears to be the neckline region.

\begin{figure}[t]
    \begin{center}
    \includegraphics[width=\textwidth]{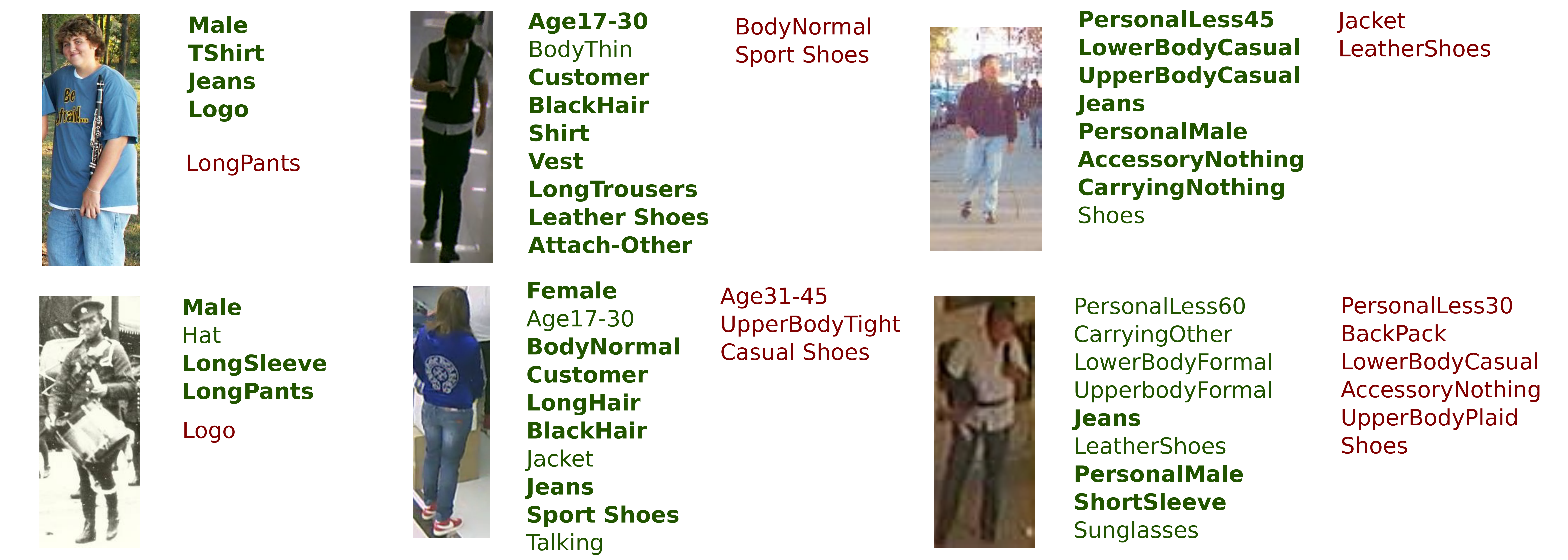}
    \end{center}
    \caption{Qualitative results of VeSPA on WIDER (left), RAP (middle) and PETA (right). Correct
    attribute predictions are marked in bold green, missed attributes in green and false positive
    predictions in red. A negative example with a number of mistakes is diplayed in the lower right
    coner. Note however, that many of the mistakes are reasonable ones (e.g. predicting \emph{shoes}
    instead of \emph{leather shoes}).}
    \label{fig:qualitative}
\end{figure}

Our qualitative analysis shows that the model predictions are indeed based on meaningful attribute localizations and image context. Some example results of VeSPA are also depicted in Figure \ref{fig:qualitative}.
%\footnote{The VeSPA model and code will be made available online after review}.

\section{Conclusion}
\label{sec:conclusion}
We have presented a unified model to jointly predict the person's view and specialized view dependent attribute inference. Our results shows that our model learns a reliable view predictor which is directly transferable to other datasets. The induced view-specific information into the attribute prediction units helps learn attributes better. In comparison to the published state-of-the-art that explicitly uses body parts, image context and scene context, our results show that relatively straight forward extensions and incorporating view information has proven useful for person attribute recognition. In addition to providing convincing semantic attribute predictions the view information may also aid in specific pedestrian search and retrieval applications.

%\section*{Acknowledgements}

\nocite{*}%TODO: remove this for submission
\bibliography{references}
\newpage
\end{document}